%% file: 0Abstract.tex
\documentclass{article}

\usepackage[final]{neurips_2024}

\usepackage[utf8]{inputenc} %
\usepackage[T1]{fontenc}    %
\usepackage{hyperref}       %
\usepackage{url}            %
\usepackage{booktabs}       %
\usepackage{amsfonts}       %
\usepackage{nicefrac}       %
\usepackage{microtype}      %
\usepackage{xcolor}         %
\usepackage{graphicx}
\usepackage{amsmath}
\usepackage{color, xspace}

\usepackage{enumitem}
\usepackage[normalem]{ulem}
\usepackage{multirow}

\title{G3: An Effective and Adaptive Framework for Worldwide Geolocalization Using Large Multi-Modality Models}

\author{
    Pengyue Jia$^1$, Yiding Liu$^2$, Xiaopeng Li$^1$, \textbf{Yuhao Wang}$^1$, \textbf{Yantong Du}$^1$,\\ \textbf{Xiao Han}$^1$, \textbf{Xuetao Wei}$^3$, \textbf{Shuaiqiang Wang}$^2$, \textbf{Dawei Yin}$^2$, \textbf{Xiangyu Zhao}$^1$\thanks{Corresponding author.} \\
    $^1$City University of Hong Kong, $^2$Baidu Inc., $^3$Southern University of Science and Technology  \\
    \texttt{\{jia.pengyue,xiaopli2-c,yhwang25-c\}@my.cityu.edu.hk}, \\
    \texttt{\{liuyiding.tanh,hahahenha,shiqiang.wang\}@gmail.com}, \\
    \texttt{duyantong94@hrbeu.edu.cn,weixt@sustech.edu.cn}, \\ \texttt{yindawei@acm.org,xianzhao@cityu.edu.hk}
}

\begin{document}

\maketitle

\begin{abstract}
  Worldwide geolocalization aims to locate the precise location at the coordinate level of photos taken anywhere on the Earth. 
  It is very challenging due to 1) the difficulty of capturing subtle location-aware visual semantics, and 2) the heterogeneous geographical distribution of image data. As a result, existing studies have clear limitations when scaled to a worldwide context. They may easily confuse distant images with similar visual contents, or cannot adapt to various locations worldwide with different amounts of relevant data. 
  To resolve these limitations, we propose \textbf{G3}, a novel framework based on Retrieval-Augmented Generation (RAG). In particular, G3 consists of three steps, i.e., \textbf{G}eo-alignment, \textbf{G}eo-diversification, and \textbf{G}eo-verification to optimize both retrieval and generation phases of worldwide geolocalization. During Geo-alignment, our solution jointly learns expressive multi-modal representations for images, GPS and textual descriptions, which allows us to capture location-aware semantics for retrieving nearby images for a given query. During Geo-diversification, we leverage a prompt ensembling method that is robust to inconsistent retrieval performance for different image queries. Finally, we combine both retrieved and generated GPS candidates in Geo-verification for location prediction.
  Experiments on two well-established datasets IM2GPS3k and YFCC4k verify the superiority of G3 compared to other state-of-the-art methods. 
  Our code\footnote{https://github.com/Applied-Machine-Learning-Lab/G3} and data\footnote{https://huggingface.co/datasets/Jia-py/MP16-Pro} are available online for reproduction.

\end{abstract}

\input{1Introduction}

\input{4RelatedWork}

\input{2Methodology}
\input{3Experiments}

\input{5Conclusion}

\clearpage
\section*{Acknowledgements}

This research was partially supported by Research Impact Fund (No.R1015-23), APRC - CityU New Research Initiatives (No.9610565, Start-up Grant for New Faculty of CityU), CityU - HKIDS Early Career Research Grant (No.9360163), Hong Kong ITC Innovation and Technology Fund Midstream Research Programme for Universities Project (No.ITS/034/22MS), Hong Kong Environmental and Conservation Fund (No. 88/2022), and SIRG - CityU Strategic Interdisciplinary Research Grant (No.7020046), Huawei (Huawei Innovation Research Program), Tencent (CCF-Tencent Open Fund, Tencent Rhino-Bird Focused Research Program), Ant Group (CCF-Ant Research Fund, Ant Group Research Fund), Alibaba (CCF-Alimama Tech Kangaroo Fund (No. 2024002)), CCF-BaiChuan-Ebtech Foundation Model Fund, and Kuaishou.
\bibliography{references}{}
\bibliographystyle{plain}

\clearpage
\appendix
\input{6Appendix.tex}

\end{document}

%% file: 1Introduction.tex
\section{Introduction}

\begin{figure}[t]
    \centering
    \includegraphics[width=\linewidth]{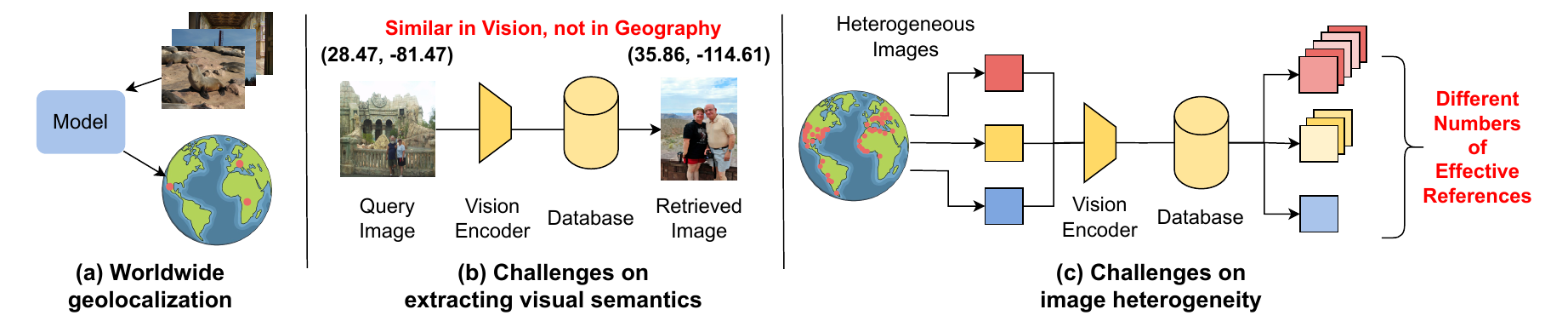}
    \caption{Illustration of limitations of prevailing methods in worldwide geolocalization.}
    \label{fig:problem}
\end{figure}

Worldwide image geolocalization~\cite{vo2017revisiting} aims to pinpoint the exact shooting location for any given photo taken anywhere on Earth, as illustrated in Figure~\ref{fig:problem}(a).
Unlike geolocalization within specific regions (e.g., at city level)~\cite{noh2017large,cao2020unifying,tan2021instance,lee2022correlation,shao2023global},  
worldwide geolocalization~\cite{cepeda2023geoclip, vo2017revisiting, zhou2024img2loc} greatly unleashes the potential of geolocalization, which is useful for various 
real-world applications, 
such as crime tracking and navigation.
However, worldwide image geolocalization is extremely challenging, as images collected from around the world are featured with a myriad of elements, including varying landscapes, weather conditions, architectural styles, etc.

Extensive research efforts have been devoted to the task, which can be broadly categorized into 1) classification-based, 2) retrieval-based, and 3) retrieval-augmented generation (RAG) methods.
In particular, classification-based methods~\cite{weyand2016planet,seo2018cplanet,pramanick2022world,clark2023we} divide the entire geographical space into fixed grids, and classify each image into a particular grid.
Retrieval-based methods convert image localization to either a 
image-to-image~\cite{cao2020unifying, lee2022correlation, shao2023global} or an image-to-GPS~\cite{cepeda2023geoclip} retrieval problem, where the final prediction is a top-retrieved GPS exists in a given candidate database. 
Generation-based methods~\cite{zhou2024img2loc} recently achieved state-of-the-art performance on this task, via applying a 
retrieval-augmented generation (RAG) pipeline that leverages the strong reasoning and generalization ability of large multi-modality models (LMMs). They usually integrate the retrieved GPS coordinates in the input prompts of LMMs as references, to generate more accurate predictions.

Despite their initial success, existing studies still have clear limitations when scaled to a worldwide context, mainly due to two challenges as shown in Figure \ref{fig:problem}.
\textbf{First}, it is very challenging to extract visual semantics that accurately indicate an image's geolocation, as two distant places could possibly have similar visual features. 
Conventional visual representations are ineffective in implying subtle location-aware semantics.
\textbf{Second}, image data usually exhibits significant heterogeneity in its geographical distribution, which existing methods can hardly handle. For retrieval-based methods, it only performs well for image queries with many nearby images stored in the database, while many images at unpopular locations may have very few or even no similar data to be compared with, leading to large prediction errors.
Worst still, such inconsistent retrieval performance significantly affects existing RAG-based methods that use a fixed number of references. As the retrieval performs inconsistently for different image queries, their generation process lacks the robustness to adapt to various image queries at different locations worldwide.

To address the aforementioned challenges, we propose G3,
a novel RAG-based solution with expressive retrieval and robust generation for worldwide image geolocalization.

\textbf{For the retrieval phase}, we train multi-modality encoders to effectively capture location-aware visual similarity between images. Unlike existing RAG-based methods that only leverage conventional visual similarity for retrieval, we propose a multi-modality alignment process, namely \textbf{Geo-alignment}, which learns the representations of GPS coordinates, images, and textual descriptions in a joint manner. By doing this, the visual representations are able to capture fine-grained location-aware semantics for retrieving other images close by. In addition to the numerical GPS coordinates, the textual descriptions (e.g., city/country names) can largely enrich the location information to be aligned with the visual representations. Moreover, to train the multi-modality representations, we establish a new dataset, namely \textbf{MP16-Pro}, by including textual geographical descriptions to the original MP16 dataset~\cite{larson2017benchmarking}. We anticipate the dataset will benefit more future work for location-aware visual representation learning.

\textbf{For the generation phase}, we leverage a prompt ensembling method, namely \textbf{Geo-diversification}, which improves the robustness of prediction generation for different types of images. More specifically, it generates a diverse set of predictions via multiple retrieval-augmented prompts, each of which might be more useful for a certain type of query images. As such, the generated GPS candidates are more likely to contain the ground truth coordinates. Subsequently, we conduct \textbf{Geo-verification}, which combines both the retrieved and the generated GPS candidates, and compares their similarities with the query image using the learned multi-modality representations. The most similar GPS is returned as the final prediction.
Extensive experiments are conducted on well-established datasets IM2GPS3k~\cite{vo2017revisiting} and YFCC4K~\cite{thomee2016yfcc100m}, and the results show the effectiveness of G3 compared to the other state-of-the-art baseline methods. 
We summarize the key contributions of our work as follows:

\begin{itemize}[leftmargin=*]
    \item We present G3, a novel solution for the worldwide geolocalization task. Our proposed method leverages 1) Geo-alignment to learn expressive location-aware representations of images, 2) Geo-diversification to improve the robustness of GPS candidate generation, and 3) Geo-verification to ensemble both retrieved and generated candidates for final prediction.
    \item We release a new dataset MP16-Pro, adding textual localization descriptions to each sample based on the original dataset MP16 to facilitate future research in the field.
    \item We extensively experiment with two well-established datasets IM2GPS3k and YFCC4K. G3 demonstrates superior performance compared to other state-of-the-art baseline methods.
\end{itemize}

%% file: 4RelatedWork.tex
\section{Related Work}

\textbf{Image Geolocalization.} Image Geolocalization is an important task in computer vision~\cite{zhu2023difftraj,zhu2023synmob,zhu2024controltraj}, spatial data mining~\cite{zhao2017incorporating,zhao2016exploring,zhang2023promptst,han2023mitigating}, and GeoAI~\cite{zhao2017modeling,zhao2022multi,zhang2023mlpst,zhang2023autostl}. Previous work in image geolocalization can be divided into three main categories: classification-based methods, retrieval-based methods, and generation-based methods. (1) Classification-based methods~\cite{seo2018cplanet,vo2017revisiting,muller2018geolocation,weyand2016planet,pramanick2022world,clark2023we} divide the entire earth into multiple grid cells and assign the center coordinates as predicted values. Models are then trained to classify the input image into the correct cell. However, if the actual location of the image is far from the center of the predicted cell, there can still be significant errors, even if the cell prediction is correct. 
(2) Retrieval-based methods treat the image geolocalization task as a retrieval problem, typically maintaining a database of images~\cite{workman2015wide,liu2019lending,zhu2021vigor,yang2021cross,zhu2022transgeo,tian2017cross,shi2020looking,zhu2023r2former} or a gallery of GPS coordinates~\cite{cepeda2023geoclip}. They take the most similar images and GPS coordinates to the query image as the predicted values. However, maintaining a global-level image database or GPS gallery is infeasible. 
(3) Generation-based methods employ large multi-modality models to generate the predicted coordinates for images~\cite{ligeoreasoner}. Zhou \textit{et al.}~\cite{zhou2024img2loc} introduced retrieval-augmented generation into the geolocalization task and took the retrieved similar images' coordinates as references to help generate predictions. However, they can not accurately extract visual semantics to indicate an image's location because they simply use visual similarity to retrieve references and suffer inaccurate prediction when facing heterogeneous query images. 
In this work, G3 introduces Geo-alignment to incorporate geographical information into image representations to help retrieve similar images in geography and proposes Geo-diversification and Geo-verification to enhance prediction performance and robustness.

\textbf{Large Multi-modality Models.}
Inspired by the success of large models~\cite{jia2018second,jia2024mill,li2023agent4ranking} in single domains like computer vision~\cite{wang2024visionllm} and natural language processing~\cite{zhao2023survey}, there has been increasing attention on large multi-modality models. CLIP~\cite{radford2021learning} aligns image and text representations through contrastive learning and achieves remarkable model generalization with simple optimization objectives. The Large Language-and-Vision Assistant (LLAVA)~\cite{liu2024visual} effectively combines CLIP's visual encoder with the powerful language model Vicuna, enhancing the model's general understanding of both visual and textual information through two-stage instruction tuning. GPT4V~\cite{achiam2023gpt} is a large multimodal model released by OpenAI in 2023, allowing users to input text and images to obtain answers.

\textbf{Retrieval-Augmented Generation.}
To mitigate the hallucination issue in LLMs, retrieval-augmented generation (RAG)~\cite{lewis2020retrieval} has emerged as a popular and effective technique. It enhances the reliability of content generated by LLMs by incorporating facts fetched from external sources. Specifically, certain factual knowledge is retrieved by a retriever from external sources based on a query. LLMs can access these retrieval results during the generation process to generate accurate outcomes. RAG preserves the generalization capabilities of LLMs while also introducing external information to enhance the reliability of generated content, efficiently alleviating the hallucination problem.

%% file: 2Methodology.tex
\section{Methodology}

\begin{figure}
    \centering
    \includegraphics[width=\linewidth]{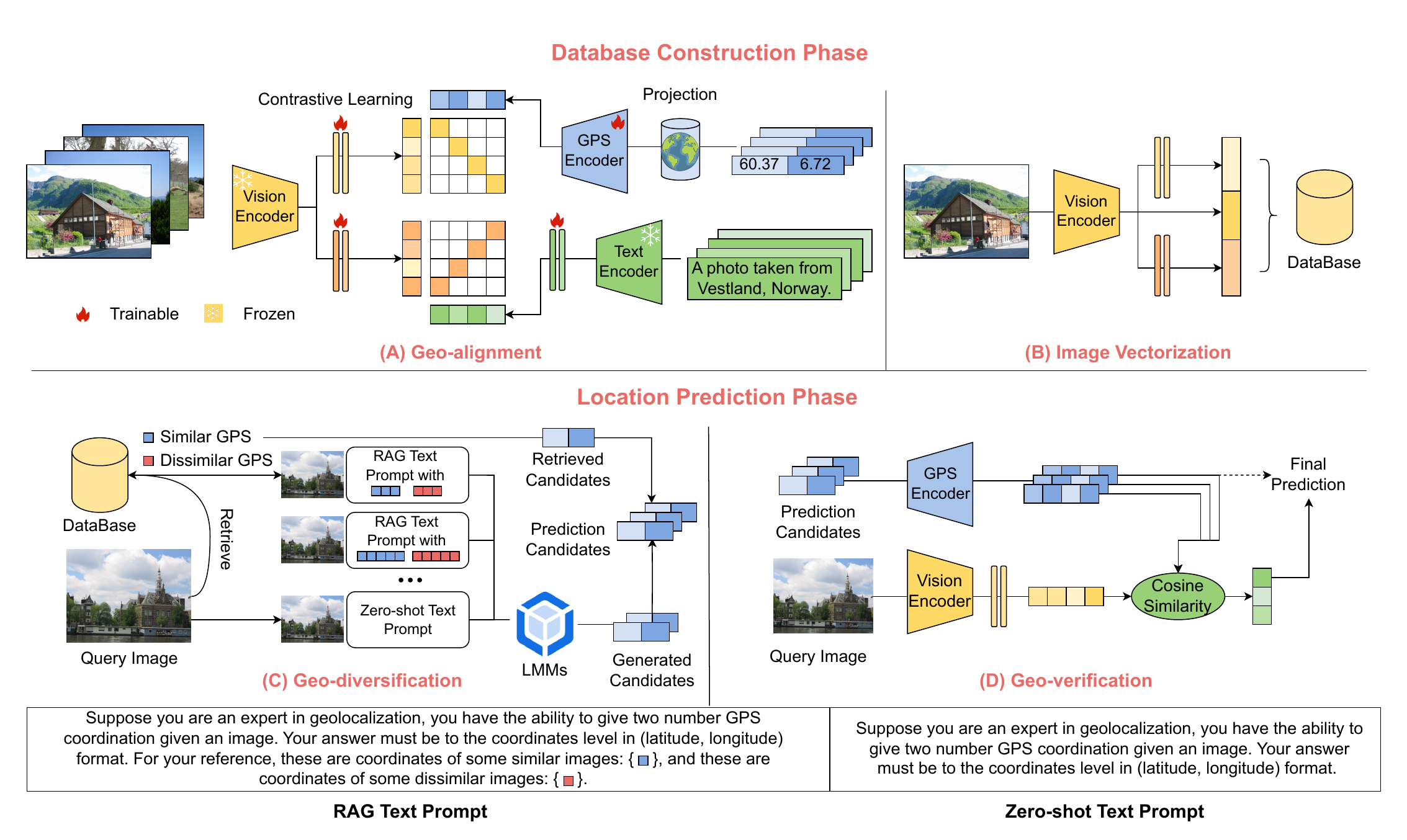}
    \caption{Overview of the framework of G3.}
    \label{fig:overview}
\end{figure}

Figure~\ref{fig:overview} illustrates the comprehensive architecture of G3, which consists of Geo-alignment, diversification, and verification, with two phases: database construction and location prediction. During the database construction phase, introduced in Section~\ref{sec:database_construction}, Geo-alignment aligns image representations with textual image descriptions and GPS information to incorporate geographical information into representations. In the location prediction phase, introduced in Section~\ref{sec:location_prediction}, similar images will be first retrieved based on the nearest neighbor search from the database; Geo-diversification will then combine their coordinates in RAG prompts to generate diverse candidates, and Geo-verification finally selects the final predicted coordinates in a multi-modality space.

\subsection{Database Construction} \label{sec:database_construction}

G3 requires an image database to preserve image representations. Existing work directly uses visual encoders (e.g., CLIP's ViT encoder or ResNet) to encode images. However, visual similarity cannot completely represent geographical proximity. To overcome this issue, we propose Geo-alignment, which incorporates geographical information into image representations by multi-modality alignment. 

\underline{\textbf{Geo-alignment.}} \label{sec:geo-alignment}
Geographical features can be divided into continuous and discrete types, which are essential in geolocalization. On the one hand, according to the first law of geography~\cite{tobler1970computer}, "everything is related to everything else, but near things are more related to each other." Climate, terrain, and vegetation are continuous features that gradually change along latitude or longitude. On the other hand, discrete features (e.g., city/country names) are also conducive to determining geographical location. These features usually change abruptly at national borders. To encode images with representations tailored for geolocalization, we propose a multi-modality alignment method, Geo-alignment, as shown in Figure~\ref{fig:overview}(A).

\textbf{Image encoding.} We use pretrained vision encoder and two trainable transformation layers to encode images:
$
    \mathbf{e}_{i,\text{text}}^{\text{image}} = f_{\text{text}}(\mathcal{V}(\mathbf{I}_i)), \ \mathbf{e}_{i,\text{gps}}^{\text{image}} = f_{\text{gps}}(\mathcal{V}(\mathbf{I}_i))
$,
where $\mathbf{e}_{i,\text{text}}^{\text{image}}$ and $\mathbf{e}_{i,\text{gps}}^{\text{image}}$ are the $i$-th image representations in the batch that need to be aligned with textual geographical descriptions and GPS data. $f_{\text{text}}$ and $f_{\text{gps}}$ are the corresponding feed forward functions, $\mathcal{V}$ represents the fixed vision encoder, and $\mathbf{I}_i$ is the $i$-th image in a batch.

\textbf{GPS coordinate encoding.} To encode GPS coordinates, an appropriate projection is needed to transform latitude and longitude into a Cartesian coordinate system. We choose not to adopt the equal earth projection (EEP) used in previous work~\cite{cepeda2023geoclip} because EEP primarily focuses on area accuracy while overlooking angular distortions, which is significant in modeling the trends of geographical features along latitude and longitude. As a result, we utilize Mercator projection for its conformal property.
The formula of Mercator projection is shown below:
\begin{align}
\begin{cases}
    \mathbf{x} = \mathbf{R} \cdot (\lambda-\lambda_0) \\
    \mathbf{y} = \mathbf{R} \cdot ln[\text{tan}(\frac{\pi}{4}+\frac{\phi}{2})]
\end{cases}
\end{align}
where $\lambda$ and $\phi$ are radians of longitude and latitude, $\lambda_0$ denotes the central meridian longitude. $\mathbf{R}$ is a proportional constant of Earth radius. The output $\mathbf{x}$ and $\mathbf{y}$ denote the transformed plane coordinates.

After projection, we follow previous work~\cite{cepeda2023geoclip} to capture high-frequency patterns and hierarchical representations using random fourier features (RFF) with various frequencies. RFF function $\gamma$ will transform the projected coordinate $\mathbf{G}_i = (\mathbf{x}_i, \ \mathbf{y}_i)$ first: $\gamma(\mathbf{G}_i) = [\text{cos}(2\pi \mathbf{M} \mathbf{G}_i), \ \text{sin}(2\pi \mathbf{S} \mathbf{G}_i)]^\text{T}$. $\mathbf{M}$ denotes a matrix sampled from a Gaussian distribution $\mathbf{M} \sim \mathcal{N}(0,\sigma)$ to limit the frequencies. To capture hierarchical representations, we sum up the outputs with different $\sigma$:
$
    \mathbf{e}_{i}^{\text{gps}} = \sum^{K}_{k=1}f_k(\gamma(\mathbf{G}_i,\  \sigma_k))
    \label{for:gps_encoding}
$
where $\mathbf{e}_{i}^{\text{gps}}$ is the encoded gps representations for $i$-th sample in a batch, $K$ denotes the number of hierarchical patterns, $f_k$ is the feed forward function for $k$-th hierarchical layer. $\sigma_k$ controls the frequency for $k$-th layer and $\sigma_k=2^{log_2(\sigma_{min}+(k-1)(log_2(\sigma_{max})-log_2(\sigma_{min}))/(N-1)},\  \forall k \in \{1, \ldots, N\}$. $\sigma_{min}$ and $\sigma_{max}$ controls the range of $\sigma_k$.

\textbf{Text encoding.} We initially employ geographical reverse encoding to obtain textual descriptions of GPS coordinates. For instance, as illustrated in Figure~\ref{fig:overview}(A), the GPS coordinates (60.37, 6.72) can be converted into the textual description "A photo taken from Vestland, Norway". These textual descriptions are inputs to a pre-trained text encoder, followed by feedforward networks for vector transformation. 
$
    \mathbf{e}_{i}^{\text{text}} = f(\mathcal{T}(\mathbf{T}_i))
$
where $\mathbf{e}_{i}^{\text{text}}$ denotes the encoded textual descriptions for $i$-th sample in a batch, $f$ is the feed forward transformation layer, $\mathcal{T}$ is the text encoder function, and $\mathbf{T}_i$ is the textual descriptions for $i$-th sample in a batch.

\textbf{Optimization.}
Geo-alignment is optimized with the following objective to align image representations with textual descriptions and GPS information:
\begin{align}
    \mathcal{L}_{a,b}=-\sum^{n}_{i=1}log(\frac{\text{exp}(\text{logits}_{ii})}{\sum_{j=1}^{n}\text{exp}(\text{logits}_{ij})}), \ \text{logits}=(\frac{\mathbf{e}^a}{\|\mathbf{e}^a\|_2})(\frac{\mathbf{e}^b}{\|\mathbf{e}^b\|_2})^\text{T} \cdot \text{exp}^{t_{a,b}}
    \label{for:loss}
\end{align}
where $\mathcal{L}_{a,b}$ denotes the loss function of modality $a$ to modality $b$, $\mathbf{e}$ is the encoded representations, and $t$ is the temperature. 
G3 needs to align image representations with both textual descriptions and GPS data, so the final optimization objective is shown below:
$
    \mathcal{L} = (\mathcal{L}_{\text{image},\text{text}} + \mathcal{L}_{\text{image},\text{gps}} + \mathcal{L}_{\text{text},\text{image}} + \mathcal{L}_{\text{gps},\text{image}}) / 2
$.

\textbf{Image vectorization.}
As illustrated in Figure~\ref{fig:overview}(B), after Geo-alignment, we will vectorize the images in the dataset and store them in a database. To maintain image representations tailored for geolocation tasks, we concatenate the original visual representations with representations aligned with geographical information: $\mathbf{e}^{\prime} = \text{concat}(\mathbf{e}^{\text{image}},\mathbf{e}^{\text{image}}_{\text{text}},\mathbf{e}^{\text{image}}_{\text{gps}})$. $\mathbf{e}^{\prime}$ denotes the final representation, $\mathbf{e}^{\text{image}}$ represents the vector obtained directly through the pretrained vision encoder. $\mathbf{e}^{\text{image}}_{\text{text}}$ and $\mathbf{e}^{\text{image}}_{\text{gps}}$ are the image representations aligned with textual geographical descriptions and GPS information.

\subsection{Location Prediction} \label{sec:location_prediction}

Figure~\ref{fig:overview}(C) and (D) illustrate the overview of the location prediction phase. Previous work~\cite{zhou2024img2loc} directly incorporates the GPS coordinates of similar retrieved images as references into a single RAG prompt to generate predictions.
However, due to the heterogeneity of query images, the number of reference GPS coordinates varies when each sample achieves optimal prediction performance.
To address this issue, Geo-diversification expands the candidate pool with prompts containing different numbers of reference coordinates, including zero (i.e., in a zero-shot manner), as shown in Figure~\ref{fig:overview}(C). Illustrated in Figure~\ref{fig:overview}(D), Geo-verification selects the best prediction coordinate for each sample using the well-trained Image-to-GPS encoders in Geo-alignment. In the location prediction phase, Geo-diversification and Geo-verification are introduced to enrich the diversity of generated predictions and select the predictions with the highest confidence. 

\underline{\textbf{Geo-diversification.}} \label{sec:geo-consistency}
Due to the heterogeneity of query images, the number of reference coordinates introduced in the RAG process varies when each sample achieves optimal prediction performance. To solve this issue, we introduce Geo-diversification. Specifically, we first construct $K$ RAG prompts with different numbers of reference coordinates (0 reference coordinates equals zero-shot generation), and each prompt will generate $N$ results. This process can be represented as follows:
$
    \{c_1^k,c_2^k,\cdots,c_n^k\} = \text{RAG}(p^k)
$
where $c^k$ denotes the candidate coordinate generated by the $k$-th RAG prompt, and $p^k$ is the $k$-th RAG prompt.
The final candidate pool contains the top $S$ coordinate candidates of retrieved similar images and the generated coordinate candidates as shown in Figure~\ref{fig:overview}. The final candidate pool is denoted as $\{c_1,c_2,\cdots,c_m\}$, where $m = K \times N + S$.

\underline{\textbf{Geo-verification.}} \label{sec:geo-verification}
Given the coordinate candidates set $\{c_1,c_2,\cdots,c_m\}$, selecting the best guess is essential and challenging. We reinvent the well-trained Image-to-GPS model in Geo-alignment to achieve this target, as shown in Figure~\ref{fig:overview}(D). The similarity between image representations $\mathbf{e}_{\text{gps}}^{\text{image}}$ and GPS representations $\mathbf{e}^{gps}$ are derived by $\text{sim} = \mathbf{e}_{\text{gps}}^{\text{image}} (\mathbf{e}^{gps})^{\text{T}}$, and the coordinate $c_j$ with the highest similarity is selected as the final prediction by $j=\text{argmax}(\text{sim}_j), \ j \in \{1,2,\cdots,m\}$.

\section{MP16-Pro Dataset}

To facilitate subsequent research, we propose the MP16-Pro dataset by adding textual geographical descriptions to each sample from the MediaEval Placing Tasks 2016 (MP-16) dataset~\cite{larson2017benchmarking}. Specifically, we utilize the open-source geocoding tool Nominatim to obtain multi-level geographical textual descriptions for each sample's GPS location (\textbf{total 4.72 million locations}). There are eight geographical unit levels: neighborhood, city, county, state, region, country, country code, and continent. Some examples are given in Appendix~\ref{sec:mp16-pro-samples} for reference. Geographical text descriptions provide additional information for geolocalization tasks and enable models to transcend the original paradigm solely supporting image and GPS alignment, facilitating more diverse modeling approaches. 

%% file: 3Experiments.tex
\section{Experiments}

\textbf{Datasets and evaluation metrics:} For database construction and model training, we use the MP16-Pro dataset we released. It contains 4.72 million geotagged images from Flickr~\footnote{https://www.flickr.com/}. However, given that the dataset was released in 2016, currently, 4.12 million images within the dataset remain accessible. Following previous work~\cite{cepeda2023geoclip,zhou2024img2loc}, we evaluate G3 with public datasets (IM2GPS3k~\cite{hays2008im2gps} and YFCC4K~\cite{thomee2016yfcc100m}) and a threshold metric. Given the predicted coordinates and the ground truths, this metric quantifies the percentage of predictions where the distance to the ground truth falls within specified thresholds (1km, 25km, 200km, 750km, and 2500km).

\textbf{Implementation details:} \label{sec:imp_details}We use faiss~\cite{douze2024faiss} to deploy the image database. The vision and text encoders are pretrained ViT-L/14 and a masked self-attention transformer from CLIP~\cite{radford2021learning}. The dimensions for two trainable layers of $f_{\text{text}}$, $f_{\text{gps}}$, $f$ are 768 and 768. The input dimension of GPS encoder is 512, and the dimensions for four hidden layers of $f_k$ in Equation~\ref{for:gps_encoding} are 1024, the output dimension is 512. For the Earth radius, we set it as 6378137.0. For RFF, we use three hierarchies with $\sigma_{min}$ as $2^0$ and $\sigma_{max}$ as $2^8$. GPT4V~\footnote{https://openai.com/} is selected as the LMMs in this paper. Its temperature is set to 1.2. The number of RAG prompts $K$ is set to 4, and the number of candidates for each RAG $N$ is set to 5 for IM2GPS3K and 1 for YFCC4K. The number of similar image coordinates taken into account in candidates is 0 for IM2GPS3K and 1 for YFCC4K. G3 is trained using AdamW optimizer with learning rate 3e-5 and weight decay 1e-6. A step linear scheduler is employed with gamma 0.87, and the training epoch is set to 10. Training batch size is set to 256 and temperature $t$ in Equation~\ref{for:loss} is initialized as 3.99. All experiments are conducted with Pytorch and one NVIDIA H800 GPU.
Please refer to Appendix~\ref{sec:more_hyper} for more details on the training environment, training time, and API cost. We also mention the limitations of G3 in Appendix~\ref{sec:limitations}.

\textbf{Baselines:} To evaluate G3 in geolocalization, we follow previous work~\cite{cepeda2023geoclip,zhou2024img2loc} and select the following baselines for comparison: [L]kNN,$\sigma$=4~\cite{vo2017revisiting}, PlaNet~\cite{weyand2016planet}, CPlaNet~\cite{seo2018cplanet}, ISNs~\cite{muller2018geolocation}, Translocator~\cite{pramanick2022world}, GeoDecoder~\cite{clark2023we}, GeoCLIP~\cite{cepeda2023geoclip}, Img2Loc~\cite{zhou2024img2loc}, PIGEON~\cite{haas2024pigeon}. The detailed descriptions of baselines are in Appendix~\ref{sec:baseline}. Due to the lack of available implementations for Img2Loc, we reproduce it based on its paper and release it in our repository for future research.

\subsection{Comparison with State-of-the-art Methods}

\begin{table}
\centering
\caption{Overall experimental results on IM2GPS3K and YFCC4K.}
\label{tab:overall}
\resizebox{\linewidth}{!}{
\begin{tabular}{ccccccccccc} 
\toprule
\multirow{2}{*}{Methods} & \multicolumn{5}{c}{IM2GPS3K}                                                                                                                                                                                                                                                          & \multicolumn{5}{c}{YFCC4K}                                                                                                                                                                                                                                                             \\ 
\cmidrule[\heavyrulewidth]{2-11}
                         & \begin{tabular}[c]{@{}c@{}}Street\\1km\end{tabular} & \begin{tabular}[c]{@{}c@{}}City\\25km\end{tabular} & \begin{tabular}[c]{@{}c@{}}Region\\200km\end{tabular} & \begin{tabular}[c]{@{}c@{}}Country\\750km\end{tabular} & \begin{tabular}[c]{@{}c@{}}Continent\\2500km\end{tabular} & \begin{tabular}[c]{@{}c@{}}Street\\1km\end{tabular} & \begin{tabular}[c]{@{}c@{}}City\\25km\end{tabular} & \begin{tabular}[c]{@{}c@{}}Region\\200km\end{tabular} & \begin{tabular}[c]{@{}c@{}}Country\\750km\end{tabular} & \begin{tabular}[c]{@{}c@{}}Continent\\2500km\end{tabular}  \\ 
\midrule
\text{[L]kNN}, sigma=4~\cite{vo2017revisiting}           & 7.2                                                 & 19.4                                               & 26.9                                                  & 38.9                                                   & 55.9                                                      & 2.3                                                 & 5.7                                                & 11                                                    & 23.5                                                   & 42                                                         \\
PlaNet~\cite{seo2018cplanet}                   & 8.5                                                 & 24.8                                               & 34.3                                                  & 48.4                                                   & 64.6                                                      & 5.6                                                 & 14.3                                               & 22.2                                                  & 36.4                                                   & 55.8                                                       \\
CPlaNet~\cite{seo2018cplanet}                  & 10.2                                                & 26.5                                               & 34.6                                                  & 48.6                                                   & 64.6                                                      & 7.9                                                 & 14.8                                               & 21.9                                                  & 36.4                                                   & 55.5                                                       \\
ISNs~\cite{muller2018geolocation}                     & 10.5                                                & 28                                                 & 36.6                                                  & 49.7                                                   & 66                                                        & 6.5                                                 & 16.2                                               & 23.8                                                  & 37.4                                                   & 55                                                         \\
Translocator~\cite{pramanick2022world}             & 11.8                                                & 31.1                                               & 46.7                                                  & 58.9                                                   & 80.1                                                      & 8.4                                                 & 18.6                                               & 27                                                    & 41.1                                                   & 60.4                                                       \\
GeoDecoder~\cite{clark2023we}               & 12.8                                                & 33.5                                               & 45.9                                                  & 61                                                     & 76.1                                                      & 10.3                                                & 24.4                                               & 33.9                                                  & 50                                                     & 68.7                                                       \\
GeoCLIP~\cite{cepeda2023geoclip}                  & 14.11                                               & 34.47                                              & 50.65                                                 & 69.67                                                  & 83.82                                             & 9.59                                                & 19.31                                              & 32.63                                                 & 55                                                     & 74.69                                              \\
Img2Loc~\cite{zhou2024img2loc}                  & \uline{15.34}                                       & \uline{39.83}                                      & 53.59                                         & 69.7                                           & 82.78                                                     & \uline{19.78}                                       & \uline{30.71}                                      & \uline{41.4}                                          & 58.11                                          & 74.07                                                      \\ 
PIGEON~\cite{haas2024pigeon} & 11.3 & 36.7 & \uline{53.8} & \textbf{72.4} & \textbf{85.3} & 10.4 & 23.7 & 40.6 & \uline{62.2} & \uline{77.7} \\
\midrule
Ours                     & \textbf{16.65}                                      & \textbf{40.94}                                     & \textbf{55.56}                                        & \uline{71.24}                                         & \uline{84.68}                                            & \textbf{23.99}                                      & \textbf{35.89}                                     & \textbf{46.98}                                        & \textbf{64.26}                                         & \textbf{78.15}                                             \\
\bottomrule
\end{tabular}
}
\end{table}

To verify the effectiveness of G3, we conduct comparative experiments on IM2GPS3K and YFCC4K with other state-of-the-art methods. The results are shown in Table~\ref{tab:overall}. (1) G3 is superior to all the other baselines on almost all metrics. In addition, compared to the second best methods, G3 achieves \textbf{\underline{8.5\%}}, \textbf{\underline{2.8\%}}, \textbf{\underline{3.3\%}} improvements on IM2GPS3K in the 1km, 25km, 200km thresholds and \textbf{\underline{21.3\%}}, \textbf{\underline{16.9\%}}, \textbf{\underline{13.5\%}}, \textbf{\underline{3.3\%}}, \textbf{\underline{0.6\%}} improvements on YFCC4K in the 1km, 25km, 200km, 750km, 2500km thresholds. (2) G3, Img2Loc, GeoCLIP, and PIGEON achieve leading results, which can be attributed to the other methods taking the worldwide geolocalization task as a classification problem, introducing inevitable systemic biases. (3) G3 demonstrates significant improvements over GeoCLIP because GeoCLIP is constrained by the settings of the GPS gallery, which can not cover the entire globe. (4) Compared to Img2Loc, G3, through Geo-alignment that aligns images with discrete and continuous geographical features, achieves more precise retrieval of reference coordinates for subsequent RAG processes. Additionally, Geo-diversification and Geo-verification effectively expand the candidate pool and filter out the confident prediction results, further enhancing geolocalization performance. Overall, G3 achieves the best performance on all datasets across almost all metrics, which verifies the effectiveness of G3.

\subsection{Ablation Study}

\begin{table}
\centering
\caption{Ablation study on IM2GPS3K and YFCC4K.}
\label{tab:ablation}
\resizebox{\linewidth}{!}{
\begin{tabular}{ccccccccccc} 
\toprule
\multirow{2}{*}{Methods} & \multicolumn{5}{c}{IM2GPS3K}                                                                                                                                                                                                                                                          & \multicolumn{5}{c}{YFCC4K}                                                                                                                                                                                                                                                             \\ 
\cmidrule{2-11}
                         & \begin{tabular}[c]{@{}c@{}}Street\\1km\end{tabular} & \begin{tabular}[c]{@{}c@{}}City\\25km\end{tabular} & \begin{tabular}[c]{@{}c@{}}Region\\200km\end{tabular} & \begin{tabular}[c]{@{}c@{}}Country\\750km\end{tabular} & \begin{tabular}[c]{@{}c@{}}Continent\\2500km\end{tabular} & \begin{tabular}[c]{@{}c@{}}Street\\1km\end{tabular} & \begin{tabular}[c]{@{}c@{}}City\\25km\end{tabular} & \begin{tabular}[c]{@{}c@{}}Region\\200km\end{tabular} & \begin{tabular}[c]{@{}c@{}}Country\\750km\end{tabular} & \begin{tabular}[c]{@{}c@{}}Continent\\2500km\end{tabular}  \\ 
\midrule
w/o Geo-A                & 15.71                                               & \uline{40.64}                                      & \uline{54.85}                                         & \uline{70.8}                                           & \uline{84.05}                                             & \uline{20.8}                                        & \uline{32.72}                                      & \uline{44.25}                                         & \uline{61.83}                                          & \uline{76.64}                                              \\
w/o Geo-D                & \uline{16.35}                                       & 40.51                                              & 53.89                                                 & 69.2                                                   & 83.11                                                     & 20.28                                               & 31.87                                              & 43.67                                                 & 60.84                                                  & 76.25                                                      \\
w/o Geo-V                & 14.98                                               & 38.27                                              & 51.25                                                 & 67.6                                                   & 81.18                                                     & 19.03                                               & 30.24                                              & 40.93                                                 & 57.93                                                  & 72.83                                                      \\ 
\midrule
Ours                     & \textbf{16.65}                                      & \textbf{40.94}                                     & \textbf{55.56}                                        & \textbf{71.24}                                         & \textbf{84.68}                                            & \textbf{23.99}                                      & \textbf{35.89}                                     & \textbf{46.98}                                        & \textbf{64.26}                                         & \textbf{78.15}                                             \\
\bottomrule
\end{tabular}
}
\end{table}

To understand the specific effects of each module in G3, we design the following variants:
\begin{itemize}[leftmargin=*]
    \item \textbf{w/o Geo-A:} G3 without Geo-alignment. Directly using  ViT in CLIP for database construction.
    \item \textbf{w/o Geo-D:} G3 without Geo-diversification. Generating prediction with one RAG prompt with 10 positive samples and 10 negative samples (the parameter has been tuned).
    \item \textbf{w/o Geo-V:} G3 without Geo-verification. Instead of using the well-trained Image-GPS model in Geo-alignment, this variant randomly selects the final prediction from candidates.
\end{itemize}
Table~\ref{tab:ablation} shows the experimental results. We can draw the following conclusions: (1) All three modules significantly contribute to the final performance. (2) G3 achieves better performance than w/o Geo-A for Geo-alignment incorporates geographical information into image representations. As a result, the retrieved images are geographically similar to the query image, enhancing the effectiveness of references in the RAG process. (3) G3 is superior to w/o Geo-D, for the number of reference coordinates varies when each sample achieves the optimal prediction performance facing heterogeneous query images. The absence of Geo-diversification leads to suboptimal candidates. (4) Comparing G3 with w/o Geo-V, we observe a significant performance drop in w/o Geo-V, indicating the necessity of Geo-verification. 

\subsection{Hyperparameter Analysis}

In the generation process within G3, two hyperparameters directly impact the results: the number of RAG prompts and the number of candidate coordinates generated by each single prompt.

\begin{figure}
    \centering
    \includegraphics[width=\linewidth]{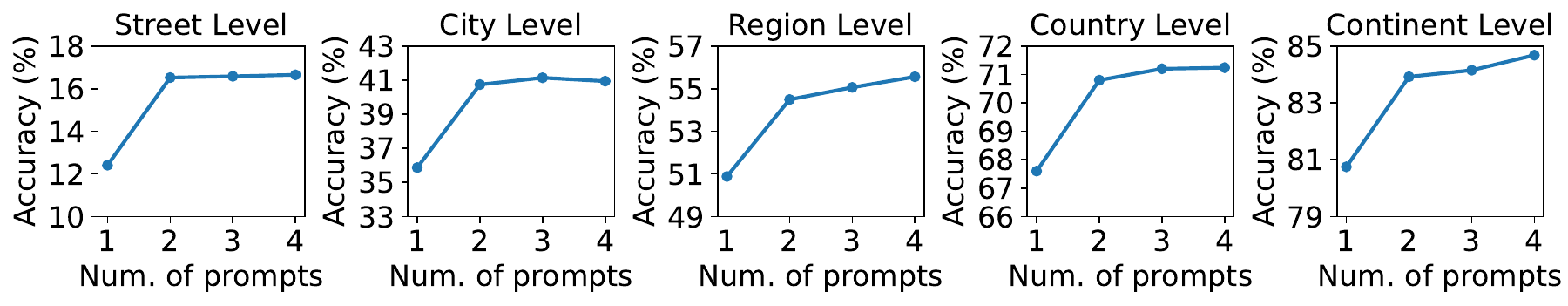}
    \caption{Varying the number of RAG prompts on IM2GPS3K.}
    \label{fig:hyper-1}
\end{figure}

\textbf{Number of RAG prompts.} To investigate the impact of varying numbers of RAG prompts, we design the following experiment: We employ four sets of RAG prompts with different reference coordinates: 0 positive, 0 negative; 5 positive, 5 negative; 10 positive, 10 negative; 15 positive, 15 negative. Starting with the first prompt, subsequent prompts will be sequentially added to change the number of RAG prompts. The number of candidates generated by each prompt is fixed to 5. As illustrated in Figure~\ref{fig:hyper-1}, the influence of RAG prompt counts on prediction performance is consistent across different metric thresholds. A significant enhancement is observed when the number increases from 1 to 2. The reason is that the zero-shot prompt (RAG prompt with 0 positive and 0 negative reference coordinates) fails to provide high-quality predictions for global images with insufficient information. The model's performance gradually improves as the number increases from 2 to 4 because having more candidates will increase the possibility of containing the ground truth coordinates. 

\begin{figure}
    \centering
    \includegraphics[width=\linewidth]{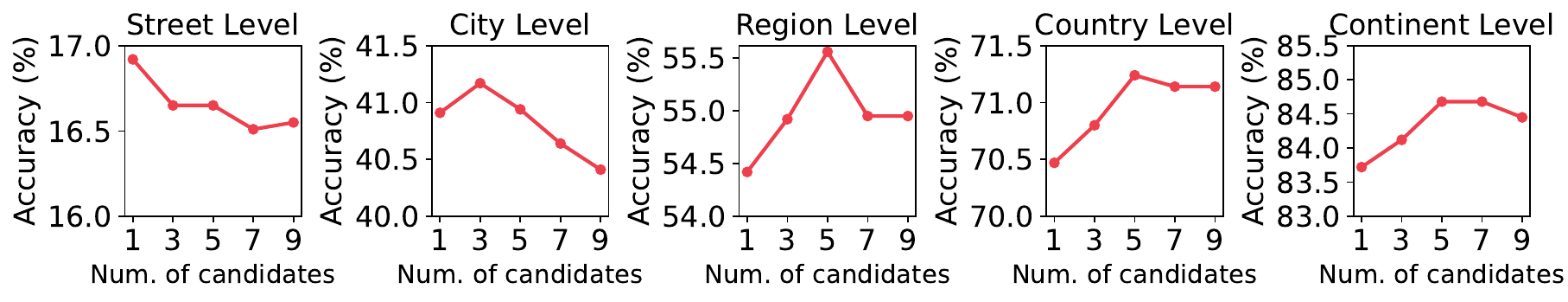}
    \caption{Varying the number of candidates for each RAG prompt on IM2GPS3K.}
    \label{fig:hyper-2}
\end{figure}

\textbf{Number of candidates.} Figure~\ref{fig:hyper-2} shows the results varying the number of candidates for each prompt. We fix the number of prompts to 4 in this experiment. We observe that the turning points where performance begins to decline exhibit an increasing trend at different levels. Specifically, at the street level, performance declines after just one candidate, at the city level after three, at the region and country levels after five, and at the continent level after seven. Three key points merit attention: (1) The initial upward trend occurs because the generation of LMMs involves randomness. Introducing more candidates can alleviate the randomness. (2) As the number of candidates increases, performance ultimately drops, likely due to the introduction of more noise in the predictions from additional generations. (3) The turning points of decline differ by level because broader levels demonstrate greater tolerance to predictive bias when more noise candidates are included.

\subsection{Effectiveness of Geo-alignment and Mercator Projection}

\begin{table}[t]
\centering
\caption{Distance statistics of retrieval reference images with different embedding methods. Avg., Md., Max., and Min. are the average, median, maximum, and minimum distances to the query image.}
\label{tab:exp4}
\resizebox{\linewidth}{!}{
\begin{tabular}{ccccccccccccc} 
\toprule
\multirow{2}{*}{Methods} & \multicolumn{4}{c}{Top-5 Candidates}                                 & \multicolumn{4}{c}{Top-10 Candidates}                                & \multicolumn{4}{c}{Top-15 Candidates}                                 \\
                                   & Avg.            & Md.             & Max.            & Min.           & Avg.            & Md.             & Max.            & Min.           & Avg.            & Md.             & Max.            & Min.            \\ 
\midrule
CLIP ViT                           & 2554.7          & 2244.6          & 5048.4          & 800.7          & 2645.9          & 2269.9          & 6376.9          & 513.3          & 2704.1          & 2307.7          & 7142.2          & 404.8           \\
G3+EEP                             & \uline{2361.5}  & \uline{2089.0}  & \uline{4618.2}  & \uline{735.3}  & \uline{2434.5}  & \uline{2102.2}  & \uline{5808.9}  & \textbf{479.3} & \uline{2464.3}  & \uline{2104.3}  & \uline{6529.2}  & \textbf{369.6}  \\
G3+Mercator                        & \textbf{2299.2} & \textbf{2054.9} & \textbf{4474.7} & \textbf{699.0} & \textbf{2362.5} & \textbf{2035.4} & \textbf{5569.6} & \uline{482.2}  & \textbf{2405.0} & \textbf{2046.9} & \textbf{6341.0} & \uline{373.0}   \\
\bottomrule
\end{tabular}
}
\end{table}

To assess the effectiveness of Geo-alignment and Mercator projection, we conduct the following experiments focusing on the reference retrieval phase: We build image databases using different embedding techniques and then retrieve the Top-N images closest to the query image. The geodesic distances will be calculated between their coordinates and the query image. The embedding variants are illustrated as follows:
\begin{itemize}[leftmargin=*]
    \item \textbf{CLIP ViT:} Directly using the visual encoder ViT in CLIP for image embedding. 
    \item \textbf{G3+EEP:} Geo-alignment with Equal Earth Projection (EEP).
    \item \textbf{G3+Mercator:} Geo-alignment with Mercator Projection.
\end{itemize}
Table~\ref{tab:exp4} shows the statistics of the geodesic distances of retrieval reference images with different embedding methods. We can draw the following conclusions: (1) G3+EEP outperforms CLIP ViT as the latter only considers visual similarity, while image representations in G3+EEP encompass both visual and geographical similarity, which is essential for geolocalization tasks. (2) G3+Mercator performs better than G3+EEP, as the EEP projection method emphasizes area projection accuracy while overlooking angular distortions, which increases the training complexity and limits the performance.

\subsection{Impact of LMMs on G3}

\begin{table}
\centering
\caption{Impact of LMMs on G3.}
\label{tab:LMMs}
\resizebox{0.9\linewidth}{!}{
\begin{tabular}{lccccc} 
\toprule
Methods         & Street 1km     & City 25km      & Region 200km   & Country 750km  & Continent 2500km  \\ 
\midrule
Img2Loc (LLaVA) & 10.21          & 29.06          & 39.51          & 56.36          & 71.07             \\
Img2Loc (GPT4V) & \uline{15.34}  & \uline{39.83}  & \uline{53.59}  & \uline{69.70}  & \uline{82.78}     \\
G3 (LLaVA)      & 14.31          & 35.87          & 49.42          & 66.93          & 81.78             \\
G3 (GPT4V)      & \textbf{16.65} & \textbf{40.94} & \textbf{55.56} & \textbf{71.24} & \textbf{84.68}    \\
\bottomrule
\end{tabular}}
\end{table}

To explore the impact of LMMs on G3, we conduct the experiments of G3 and Img2Loc with LLaVA (LLaVA-Next-LLaMA3-8b) on IM2GPS3K.
From Table~\ref{tab:LMMs} we can find that: (1) After switching LMMs from GPT4V to LLaVA, the performance of G3 shows some decline across various metrics but remains competitive. (2) Additionally, compared to Img2Loc (LLaVA), G3 (LLaVA) significantly outperforms Img2Loc (LLaVA), demonstrating the effectiveness of the proposed modules. (3) Finally, by comparing the performance of G3 equipped with LLaVA and GPT4V to Img2Loc equipped with LLaVA and GPT4V, we can observe that G3 shows more stable performance across different LMMs.

\subsection{Necessity Analysis of Three Representations Alignment in Geo-alignment}

To verify the necessity of aligning the three representations in Geo-alignment, we conduct experiments of the following variants on IM2GPS3K:
\begin{itemize}[leftmargin=*]
    \item \textbf{IMG:} Directly using pre-trained CLIP vision encoder as the encoder.
    \item \textbf{IMG+GPS:} Aligning Image representations with GPS representations in Geo-alignment, the textual descriptions are not used.
    \item \textbf{IMG+GPS+TEXT(G3):} Aligning three modalities simultaneously in Geo-alignment.
\end{itemize}

Table~\ref{tab:geo-alignment-necessity} shows that: (1) By comparing IMG+GPS+TEXT, IMG+GPS, and IMG, we find that adding GPS and text information can both enhance the feature representation compared to using the original image information alone. (2) By comparing IMG+GPS+TEXT with IMG+GPS, we find that IMG+GPS performs better at smaller scales, while IMG+GPS+TEXT performs better at larger scales. This might be because GPS is suitable for modeling variations at smaller scales, whereas text descriptions do not vary significantly at small scales and may even remain the same.

\begin{table}[h]
\centering
\caption{Results of necessity analysis.}
\label{tab:geo-alignment-necessity}
\resizebox{0.9\linewidth}{!}{
\begin{tabular}{lccccc} 
\toprule
Methods & Street 1km     & City 25km      & Region 200km   & Country 750km  & Continent 2500km  \\ 
\midrule
IMG     & 15.71          & 40.64          & 54.85          & 70.8           & 84.05             \\
IMG+GPS & \textbf{16.91} & \textbf{41.41} & \uline{55.02}  & \uline{70.94}  & \uline{84.18}     \\
IMG+GPS+TEXT(G3)      & \uline{16.65}  & \uline{40.94}  & \textbf{55.56} & \textbf{71.24} & \textbf{84.68}    \\
\bottomrule
\end{tabular}}
\end{table}

\subsection{Case Study}

\begin{figure}
    \centering
    \includegraphics[width=\linewidth]{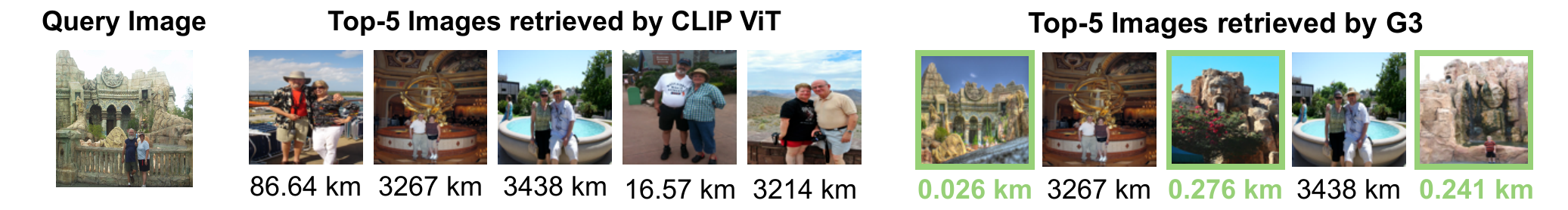}
    \caption{Reference image retrieval with CLIP ViT and G3.}
    \label{fig:case_study_1}
\end{figure}

\textbf{Case study on reference image retrieval.} Figure~\ref{fig:case_study_1} visually demonstrates the superiority of G3 in reference image retrieval. It is evident that if CLIP's ViT is used as the image encoder, the model primarily focuses on the human figures in the image (i.e., 'two people posing together in the center of the photo') while neglecting background elements beneficial for geolocalization. In Geo-alignment, G3 incorporates geographical information into the image representations. As a result, retrieved images are more focused on geographical proximity (three reference images within 1 km of the actual shooting location are retrieved in the top-5 candidate images). These valuable reference images further assist the RAG process to enhance final prediction performance.

\begin{figure}[t]
    \centering
    \includegraphics[width=\linewidth]{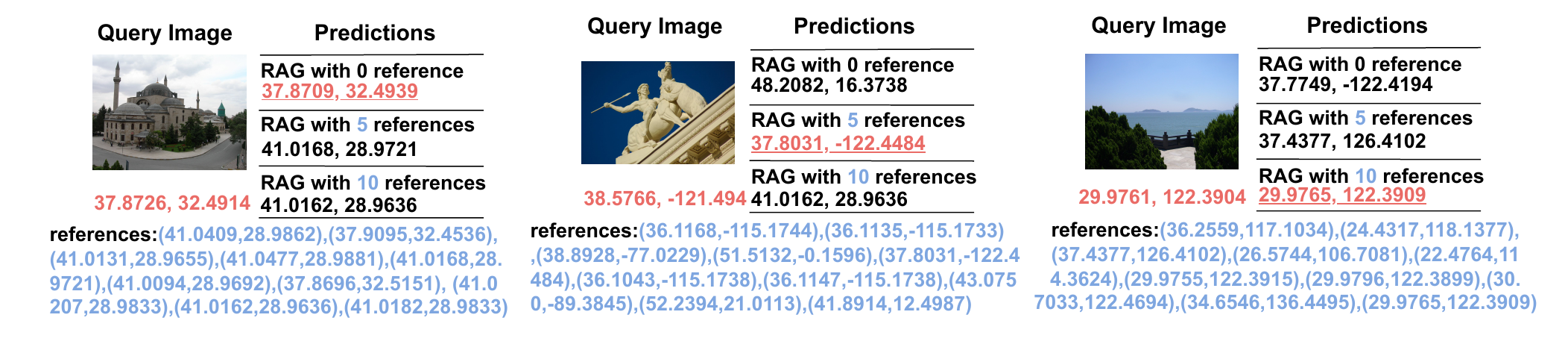}
    \caption{Predictions given with different numbers of references facing heterogeneous images.}
    \label{fig:case_study_2}
\end{figure}

\textbf{Case study on heterogeneous query image in RAG process.} Figure~\ref{fig:case_study_2} provides three examples illustrating the best prediction occurs when using RAG prompts with different numbers of references facing heterogeneous query images. (1) RAG with 0 references achieves the best performance for the first query image. This is because, on the one hand, the references are filled with biased coordinates, and on the other hand, the building in the figure is a famous landmark named Selimiye Camii mosque. The pre-trained LMMs effectively provide the longitude and latitude of this landmark based on its world knowledge. (2) For the second query image, RAG with 5 references performs best because the optimal reference appeared in the fifth position. More references do not add extra helpful information but instead introduce more noise, causing the performance of RAG with 10 references to decline; RAG with 0 references produces incorrect predictions due to the absence of clear landmark indicators in this image. (3) For the third query image, RAG with 10 references yields the best accuracy, as the references from 6 to 10 provide substantial helpful information, whereas the first five reference coordinates are far from the ground truth. Overall, from these examples, we can discern some common patterns: for images with prominent landmark features, RAG with 0 references often yields good results; for images with less informative content (such as oceans, skies, or indoor scenes), RAG with 10 references makes more comprehensive judgments based on a greater number of references; and for images with distinct regional features (images between the first two settings), RAG with 5 references will achieve satisfactory prediction accuracy.

%% file: 5Conclusion.tex
\section{Conclusion}

In this paper, we propose a novel worldwide geolocalization framework named G3. First, we introduce Geo-alignment to capture location-aware semantics in images by aligning images with textual geographical descriptions and GPS information. Second, Geo-diversification is proposed to improve the robustness of prediction generation via a prompt ensemble technique. Finally, Geo-verification selects the final coordinate prediction using the learned multi-modality representations. G3 is evaluated on two well-established datasets, IM2GPS3K and YFCC4K, and achieves state-of-the-art performance. In addition, we release a new dataset MP16-Pro, adding textual localization descriptions to each sample based on the original dataset MP16 to facilitate future research in the field. All the code and data used in this work have been released public.

%% file: 6Appendix.tex
\section{Appendix}

\subsection{MP16-Pro Dataset Samples} \label{sec:mp16-pro-samples}

To facilitate understanding of the MP16-Pro dataset, we provide some samples from the dataset in this section. The dataset contains two parts: the image data and the metadata for images.

\begin{figure}
    \centering
    \includegraphics[width=\linewidth]{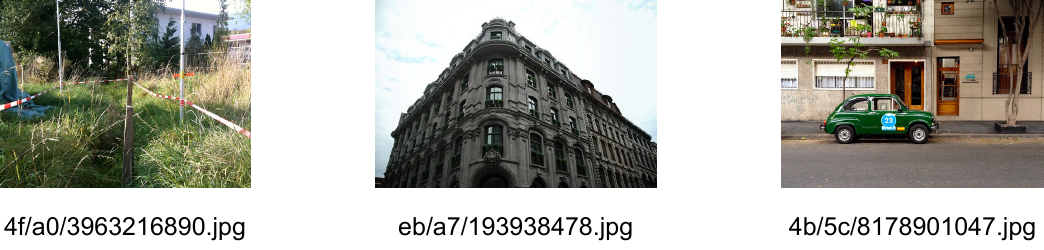}
    \caption{Image data in MP16-Pro Dataset.}
    \label{fig:mp16-pro-image}
\end{figure}

Figure~\ref{fig:mp16-pro-image} gives three examples from the MP16-Pro dataset. Take these three images as examples, MP16-Pro adds extra textual descriptions based on their coordinates: 
\begin{itemize}[leftmargin=*]
    \item For image 4f/a0/3963216890.jpg, LAT: 47.217578, LON: 7.542092, neighbourhood: Wengistein, city: Solothurn, county: Amtei Solothurn-Lebern, state: Solothurn, region: NA, country: Switzerland, country\_code: ch, continent: NA.
    \item For image eb/a7/193938478.jpg, LAT: 39.950477, LON: -75.157535, neighbourhood: Center City, city: Philadelphia, county: Philadelphia County, state: Pennsylvania, region: NA, country: United States, country\_code: us, continent: NA.
    \item For image 4b/5c/8178901047.jpg, LAT: -34.580365, LON: -58.425464, neighbourhood: Palermo, city: Buenos Aires, county: NA, state: Autonomous City of Buenos Aires, region: NA, country: Argentina, country\_code: ar, continent: NA.
\end{itemize}

All the data has been released online\footnote{https://huggingface.co/datasets/Jia-py/MP16-Pro}.

\subsection{More Details on Training and Inference} \label{sec:more_hyper}

\begin{table}
\centering
\caption{More details on training and inference parameters.}
\label{tab:more_hyper}
\resizebox{0.8\linewidth}{!}{
\begin{tabular}{cc} 
\toprule
Parameter            & G3                                                \\ 
\midrule
GPU                  & H800 80G * 1                                      \\
Training Time        & 7 hours / epoch * 10 epoch                        \\
Total params         & 441,266,179                                       \\
Trainable params     & 13,648,131 (3.09\%)                               \\
GFLOPS               & 304.38                                            \\
Dataset Samples      & 4.12M                                             \\
Batch Size           & 256                                               \\
GPU Memory           & 24G                                               \\
Text Processor       & Huggingface CLIP default text processor           \\
Vision Processor     & Huggingface CLIP default vision processor         \\
Token per RAG prompt & Input: 200+30 $\times$ k Output: 18 $\times$ n for each RAG prompt  \\
Token for IM2GPS3K   & Input: 5.1M(\$51) Output: 2.16M(\$64.8)           \\
Token for YFCC4K     & Input: 7.2M(\$72) Output: 3.06M(\$91.8)           \\
\bottomrule
\end{tabular}
}
\end{table}

In this section, we detail the training and inference process by providing information about the training environment, text and vision processors, and token costs. The Input token cost and output token cost for each RAG prompt are $200+30 \times k$ and $18 \times n$, where 200 is the fixed token cost for prompts and images with resolution parameter "low", and the 30 is related to the reference coordinates, $k$ denotes the number of references in the prompt. For the output, 18 is the token cost for one-time generation, and $n$ is the times of generations.

\subsection{G3 with Query Image}
\begin{figure}
    \centering
    \includegraphics[width=\linewidth]{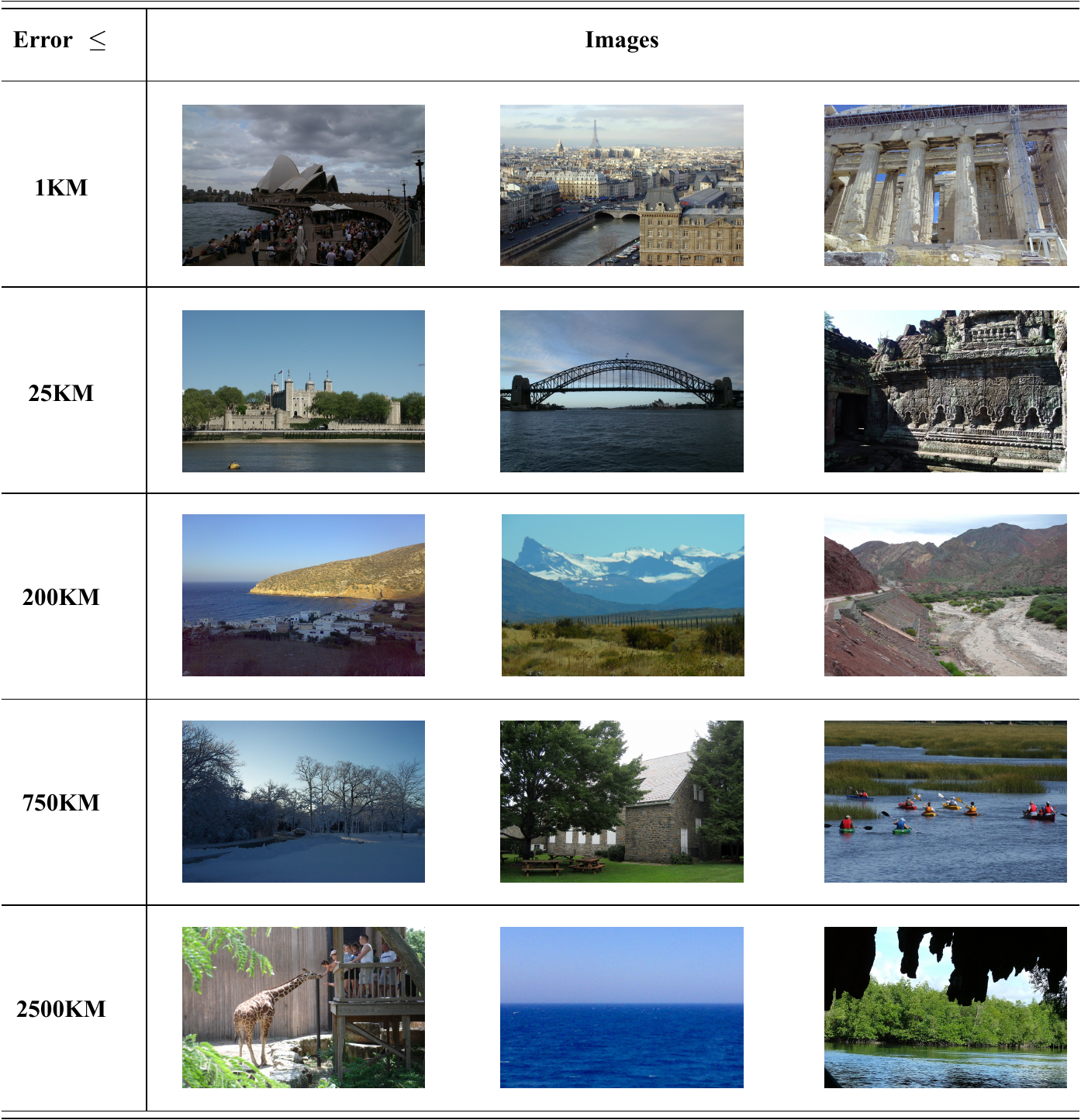}
    \caption{Example query images from IM2GPS3K that G3 localization error falls in 1km, 25km, 200km, 750km, 2500km thresholds.}
    \label{fig:gallery}
\end{figure}

Figure~\ref{fig:gallery} showcases some example query images from the IM2GPS3K dataset, with different rows representing the varying localization errors of G3 on these images. We can observe patterns in query images with errors ranging from 1km to 2500km. Images with lower errors often feature prominent landmarks or regional characteristics, such as buildings, decorations, and symbols. On one hand, these images are more likely to have more similar images in the database. On the other hand, LMMs are more sensitive to this type of information and can provide more accurate predictions. In contrast, query images with higher errors contain less effective information; large expanses of ocean, water bodies, and snow scenes offer limited assistance for geolocalization.

\subsection{Limitations} \label{sec:limitations}

The limitation of G3 is mainly about its efficiency. G3 relies on a large number of candidates generated through Geo-diversification. However, the noise candidates generated by Geo-diversification lead to low efficiency in the use of computational resources. A potential solution is to expand the database or enhance the geolocalization ability of LMMs by fine-tuning to help create prediction candidates more efficiently and effectively. In another aspect, the image vector length in Geo-alignment comprises three concatenated vectors, which increases storage demand and retrieval latency. A potential solution is to use quantizers, such as ProductQuantizer~\cite{jegou2010product}, to accelerate vector retrieval speed and reduce storage pressure.

\subsection{Details on Baselines} \label{sec:baseline}

\begin{itemize}[leftmargin=*]
    \item \textbf{\text{[L]}kNN,$\sigma=4$~\cite{vo2017revisiting}.} KNN makes use of the top k NN images and aggregates their coordinates to a prediction point. As k decreases, $\sigma$ decreases, this method transforms to NN.
    \item \textbf{PlaNet~\cite{seo2018cplanet}.} PlaNet is the first work posing worldwide geolocalization as a classification task by dividing the globe into thousands of multi-scale geographical cells. It can combine the complex clues in images to help pinpoint the shooting location of images.
    \item \textbf{CPlaNet~\cite{seo2018cplanet}.} CPlaNet tries to solve the trade-off of cell granularity. It introduces combinatorial partitioning, which creates detailed output classes by intersecting broad earth partitions, with each classifier voting for overlapping classes.
    \item \textbf{ISNs~\cite{muller2018geolocation}.} ISNs combines the hierarchical information that existed in the partitionings and the photo's scene contextual information (e.g., indoor, natural, or urban, etc.) to give the prediction.
    \item \textbf{Translocator~\cite{pramanick2022world}.} Translocator is a dual-branch transformer network that focuses on the detailed clues in images and generates robust feature representations. The semantic segmentation map and the entire image will be the input to translator.
    \item \textbf{GeoDecoder~\cite{clark2023we}.} GeoDecoder argues that previous work fails to exploit the detailed cues in different hierarchical levels. It proposes a cross-attention network to capture the complex relationships between different hierarchical features.
    \item \textbf{GeoCLIP~\cite{cepeda2023geoclip}.} GeoCLIP is based on the CLIP backbone model and first introduces a GPS encoder to transform coordinates into embeddings in worldwide geolocalization tasks.
    \item \textbf{Img2Loc~\cite{zhou2024img2loc}.} Img2Loc combines the RAG paradigm into worldwide geolocalization and is the latest work in this field. It first retrieves similar images via visual similarity and puts the coordinates of these images into RAG prompt to help generate predictions.
    \item \textbf{PIGEON~\cite{haas2024pigeon}.} PIGEON introduces an innovative approach combining semantic geocell creation, multi-task contrastive pretraining, and a novel loss function, and uniquely enhances guess accuracy through retrieval over location clusters.
\end{itemize}

\subsection{More Experimental Results on Geo-alignment.}

In this section, we conduct experiments on incorporating more fine-grained textual descriptions in Geo-alignment. Specifically, in addition to including the city, county, and country information in the textual descriptions of coordinates, we also introduce neighborhood information, which is the most fine-grained data that can be obtained from Nominatim. We use G3-N to denote this variant and keep the other hyperparameters the same as G3. The experimental results on IM2GPS3K are presented in Table~\ref{tab:granularity}.

\begin{table}
\centering
\caption{Experimental results on textual description granularity on IM2GPS3K.}
\label{tab:granularity}
\resizebox{0.6\linewidth}{!}{
\begin{tabular}{cccccc} 
\toprule
Methods & \begin{tabular}[c]{@{}c@{}}Street\\1km\end{tabular} & \begin{tabular}[c]{@{}c@{}}City\\25km\end{tabular} & \begin{tabular}[c]{@{}c@{}}Region\\200km\end{tabular} & \begin{tabular}[c]{@{}c@{}}Country\\750km\end{tabular} & \begin{tabular}[c]{@{}c@{}}Continent\\2500km\end{tabular}  \\ 
\midrule
G3-N    & 16.44                                               & 40.64                                              & 54.35                                                 & 70.57                                                  & 83.98                                                      \\
G3      & \textbf{16.65}                                      & \textbf{40.94}                                     & \textbf{55.56}                                        & \textbf{71.24}                                         & \textbf{84.68}                                             \\
\bottomrule
\end{tabular}}
\end{table}

From the results, we can see that G3 outperforms G3-N across all metrics. This may be because the text encoder's pre-training corpus contains very few instances of neighborhood-level information, resulting in weaker modeling capabilities for neighborhood names. Therefore, introducing neighborhood information into the textual descriptions of coordinates actually adds noise, which negatively impacts the effectiveness of Geo-alignment and subsequently reduces the model's prediction accuracy.

\subsection{Hard Sample and Failure Analysis}

\begin{figure}[!h]
    \centering
    \includegraphics[width=0.7\linewidth]{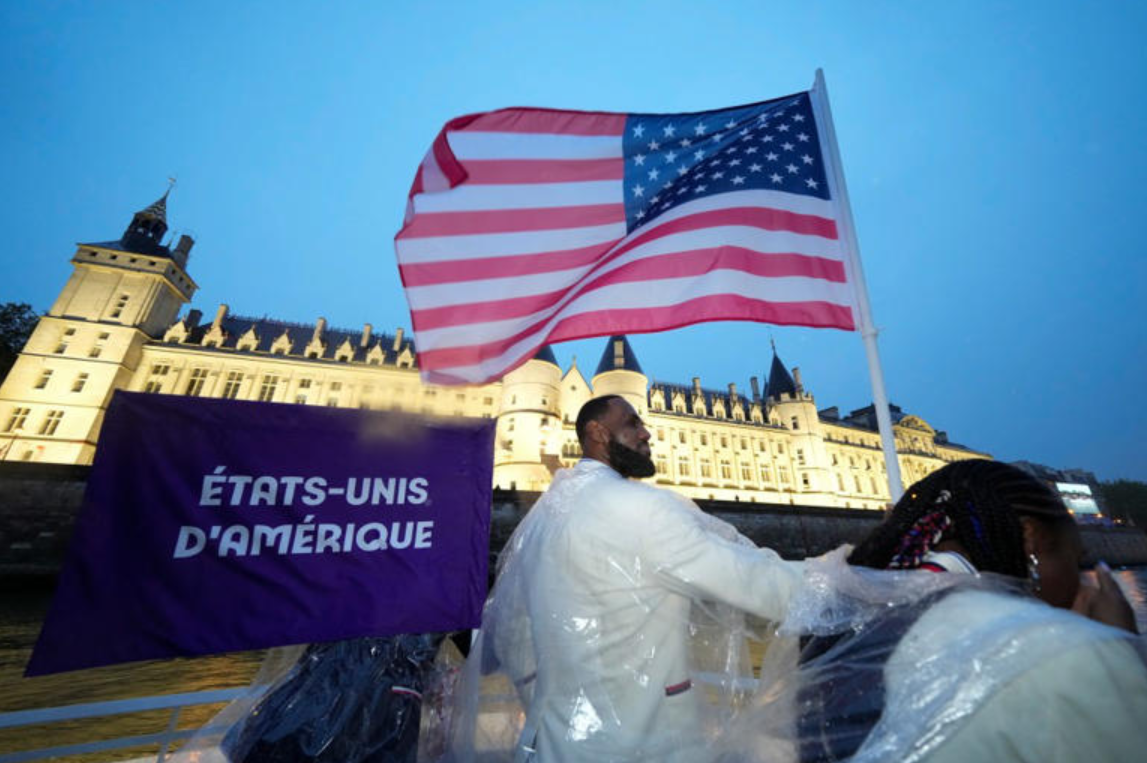}
    \caption{Hard sample.}
    \label{fig:hard_sample}
\end{figure}

\textbf{Hard Sample.} In this section, we give a hard sample to illustrate the effectiveness of G3.
Figure~\ref{fig:hard_sample} shows a man holding an American flag in France, the text on the left says ``United States of America'' in French. G3 can accurately give the prediction coordinate latitude: 48.8529 and longitude: 2.3632 located in Paris, France. This demonstrates that G3 can effectively avoid the influence of text in images, thereby focusing on the location where the image was taken, showing strong stability.

\begin{figure}[!h]
    \centering
    \includegraphics[width=\linewidth]{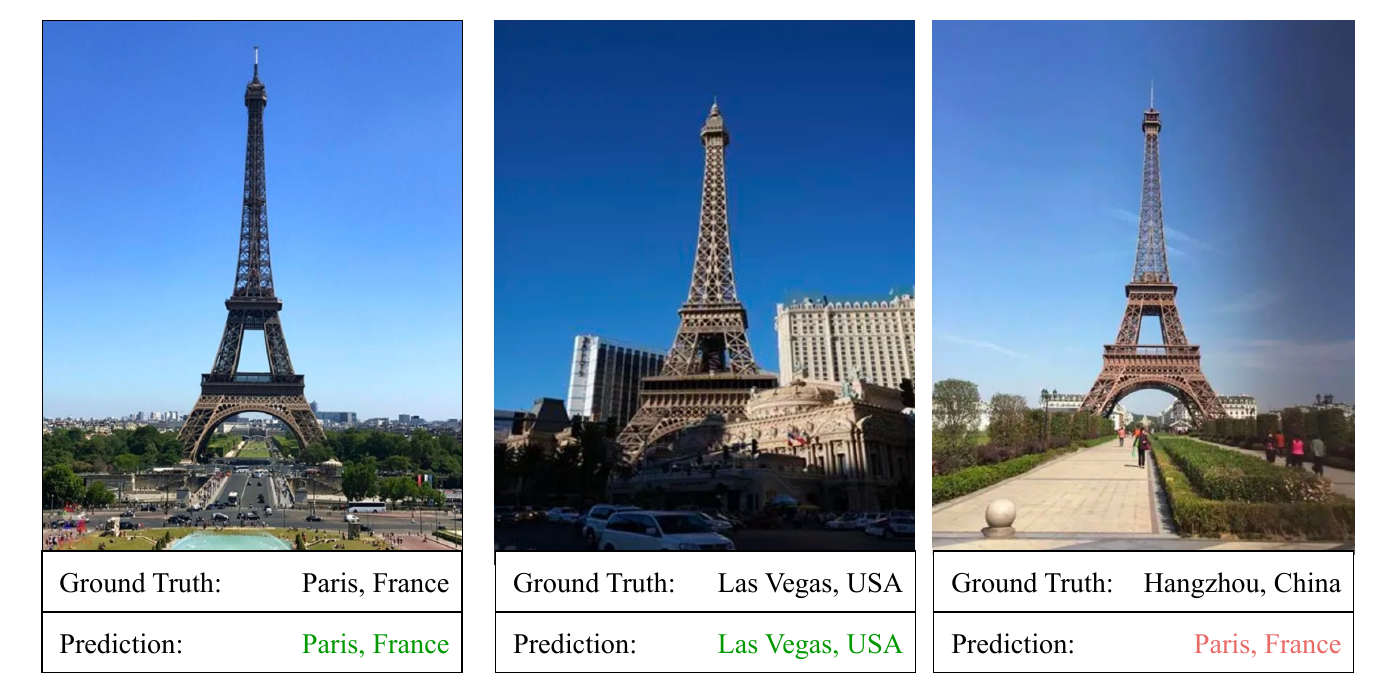}
    \caption{Experimental results of G3 on predicting coordinates of Eiffel Tower and its replica.}
    \label{fig:failure_analysis}
\end{figure}

\textbf{Failure Analysis.} Figure~\ref{fig:failure_analysis} presents the results of G3 predicting the coordinates of the Eiffel Tower and its replicas. G3 is able to locate the Eiffel Tower in France and its replica in the USA, but fails to distinguish the replica in China. The prediction for the replica in the USA is correct, while the prediction for the replica in China is incorrect. This may be because there are more reference objects around the tower in the USA, which reduces the difficulty of the model's prediction. In contrast, the lack of reference objects around the tower in China confuses G3's judgment.

